%% file: root.tex
\documentclass[letterpaper, 10 pt, conference]{ieeeconf}  

\IEEEoverridecommandlockouts                              

\title{\LARGE \bf
Using Petri Nets as an Integrated Constraint Mechanism
for Reinforcement Learning Tasks
}

\author{Timon Sachweh$^{+,1,*}$, Pierre Haritz$^{+,1,*}$ and Thomas Liebig$^{1,2,*}$
\thanks{$^{+}$ \textit{Both authors contributed to this work equally.}}%
\thanks{$^{1}$Faculty of Computer Science, Chair of Artificial Intelligence, TU Dortmund University, Dortmund, Germany}%
\thanks{$^{2}$Lamarr Institute for Machine Learning and Artificial Intelligence, Dortmund, Germany}%
\thanks{$^{*}$ \tt\small firstname.lastname@tu-dortmund.de}%
\thanks{This work was supported by the Federal Ministry of Education and Research of Germany and the state of North-Rhine Westphalia as part of the Lamarr-Institute for Machine Learning and Artificial Intelligence and by the Federal Ministry for Economic Affairs and Climate Action of Germany under grant no. 19S21005N GAIA-X 4 ROMS.}
\thanks{\textit{©2024 IEEE.  Personal use of this material is permitted.  Permission from IEEE must be obtained for all other uses, in any current or future media, including reprinting/republishing this material for advertising or promotional purposes, creating new collective works, for resale or redistribution to servers or lists, or reuse of any copyrighted component of this work in other works.}}
}

\usepackage{cite}
 
\usepackage{amsmath,amssymb,amsfonts}
\usepackage{algorithmic}
\usepackage{graphicx}
\usepackage{textcomp}
\usepackage{xcolor}
\usepackage{flushend}

\newcommand{\term}[1]{\textit{#1}}
\usepackage{algorithm2e}[ruled]
\newcommand{\nosemic}{\renewcommand{\@endalgocfline}{\relax}}%

\usepackage{comment}
\usepackage{amsthm}
\newtheorem{definition}{Definition}

\usepackage{multirow}
\usepackage{tikz}
\usetikzlibrary{arrows.meta}
\usetikzlibrary{positioning}
\usetikzlibrary{calc}

\begin{document}

\maketitle
\thispagestyle{empty}
\pagestyle{empty}


\begin{abstract}
The lack of trust in algorithms is usually an issue when using Reinforcement Learning (RL) agents for control in real-world domains such as production plants, autonomous vehicles, or traffic-related infrastructure, partly due to the lack of verifiability of the model itself.
In such scenarios, Petri nets (PNs) are often available for flowcharts or process steps, as they are versatile and standardized. In order to facilitate integration of RL models and as a step towards increasing AI trustworthiness, we propose an approach that uses PNs with three main advantages over typical RL approaches: Firstly, the agent can now easily be modeled with a combined state including both external environmental observations and agent-specific state information from a given PN. 
Secondly, we can enforce constraints for state-dependent actions through the inherent PN model.
And lastly, we can increase trustworthiness by verifying PN properties through techniques such as model checking.
We test our approach on a typical four-way intersection traffic light control setting and present our results, beating cycle-based baselines.
\end{abstract}


\section{Introduction}
Reinforcement Learning (RL) is a machine learning paradigm that includes a variety of algorithmic approaches, foremost in sequential decision-making environments. An RL model, called the agent, interacts with its feedback-providing environment in order to gain information over time to optimally act by following the learned policy. With its closed-loop properties, it is closely related to control theoretical approaches.
Since training an RL agent usually relies on a trial-and-error principle, it is less suited for applications where constraints on states are necessary. This class of RL algorithms is called Safe Reinforcement Learning.
\cite{garcia2015comprehensive, gu2023review} show various approaches that try to solve this issue mainly by changing the exploration process or by changing the optimization criterion.
Other methods, such as \textit{Shielding} \cite{alshiekh2017shielding, odriozola2023shielded}, propose a safety guard on state-specific actions, based on a safety automaton,  that triggers when actions would be chosen by the agent that lead to a critical state. The action is then simply prevented and replaced by a valid one.

One problem associated with this approach is that complete knowledge of the environment and its critical state descriptions has to be available a-priori and then need to manually be implemented via a separate shielding mechanism. Other planning-based approaches to managing macroscopic urban traffic, such as \cite{vallati2016efficient}, scale badly for the similar reasons and are ill-suited for problems where environment dynamics are unknown.

We consider cases in which descriptions of processes exist, as is typical for aforementioned business applications since standardized process models are widely used and implemented. An example of a description of a process model are Petri nets (PNs) \cite{murata1989petri}, whose properties allow them to be analyzed and validated.
These types of models can be used for traffic junctions by limiting the action space for traffic lights at each lane \cite{di2004urban, badamchizadeh2010deterministic} and additionally are able to include time constraints (\term{delays}) or use colored markings to describe more complex systems and interactions.

Traffic light control has long been a research topic for the RL community. One example is \cite{arel2010reinforcement}, where the authors consider multiple intersections in a closed system and develop a multi-agent reinforcement learning approach to control them.

In the context of process analysis, RL has been used to learn the structures of Petri net models directly. \cite{feng2010learning} use RL algorithms to learn edge connections between different places, which can then be used to insert constraints into the environment as external knowledge.

In this paper, we propose a novel approach that specifically focuses on the use of Petri nets to constrain actions on the one hand and model the agent on the other hand.
We further introduce a generic wrapper that uses constraints extracted from Petri nets to apply to RL environments. In addition, we show our idea of an RL agent, based on \textit{Deep Q~Learning}, that directly enforces constraints inside the Q-update function.
Finally, we evaluate our approach by showing results in a 4-way traffic junction environment with different reward functions and with/without the influence of constraints.

First, we will introduce PNs and RL as preliminaries, followed by a description of our approach.
Afterward, we present the results and highlight related algorithms before we come to a conclusion.

\section{Preliminaries}
In this section, we will introduce the notation and background for our problem setting.
\subsection{Petri nets}
Petri nets are graph-based models often used to model, design and analyze discrete event dynamic systems. Such systems can be event-driven, concurrent, and asynchronous. By extending standard Petri nets with firing times, it is possible to analyze its temporal performance further \cite{wang2019formal}, both in deterministic and stochastic transition cases. A Petri net \cite{murata1989petri} is defined as $PN = (G, \omega, M_0)$, where $G = (P, T, F)$ denotes the net graph with places $P$, transitions $T$ and arcs $F \subseteq (P \times T) \cup (T \times P)$, $\omega: F \rightarrow \mathbb{N_+}$ is a weight function and $M_0: P \rightarrow \mathbb{N}$ is the initial marking function that assigns sets of tokens to places.

The behavior of many systems can be described in terms of system states and their changes over time. Petri nets simulate the dynamic properties of a system and behave according to the following transition (firing) rules:
\begin{enumerate}
    \item  A transition $\bar{t} \in T$ is said to be enabled if each input place $p \in P$  is marked with at least $\omega(f)$ tokens for $f \in F$. Here, $\omega(f)$ is the weight of the arc from $p$ to $\bar{t}$.
    \item An enabled transition may or may not fire (depending on whether or not the event actually takes place) and therefore can also be triggered by a control mechanism.
    \item A firing of an enabled transition $\bar{t}$ consumes $\omega(f)$ tokens from each input place $p$ of $\bar{t}$, and creates $\omega(f')$ tokens at each output place $p$ of $t$ , where $\omega(f')$ is the weight of the arc from $\bar{t}$ to $p$.
\end{enumerate}

\subsection{Reinforcement Learning}
A Reinforcement Learning (RL) agent is typically modeled to act in sequential decision-making processes. 
\subsubsection{Markov Decision Processes}
Its environment can be formulated as a Markov Decision Process (MDP). An MDP is defined by a tuple $(\mathcal{S}, \mathcal{A}, p, \mathcal{\gamma}, r)$, where $\mathcal{S}$ denotes the state space, $\mathcal{A}$ the action space, $p: \mathcal{S} \times \mathcal{A} \rightarrow \mathcal{S}$ is the transition function and $R: \mathcal{S} \times \mathcal{A} \rightarrow \mathbb{R}$ the reward function. Additionally, a discount factor $\gamma \in [0,1)$ can be used to discount future actions. 
\subsubsection{Constrained Markov Decision Processes}
For the class of Safe RL problems, we can further introduce constraints and therefore extend the MDP by adding a constraint cost function $\mathcal{C}: \mathcal{S} \times \mathcal{A} \rightarrow \mathbb{R}$ analog to the reward function and a safety threshold $d \in \mathbb{R}$. We define a Constrained Markov Decision Process (CMDP) $M_C = (\mathcal{S}, \mathcal{A}, p, \gamma, r, \mathcal{C}, d)$. We can, in the general case, calculate a weighted return value for constrained problems with $J_C(\pi) = \underset{\tau \sim \pi}{\mathbb{E}}[\sum^\infty_{t=0} \gamma^t \mathcal{C}(s_t, a_t)]$ for a policy $\pi: \mathcal{S} \rightarrow \mathcal{A}$ with $\pi \in \Pi$ for the set of all policies $\Pi$ and a trajectory $\tau = (s_0, a_0, s_1, a_1, \dots)$.
Let $\Pi_\mathcal{C} = \{\pi \in \Pi : J_\mathcal{C}(\pi) \leq d\}$ be the set of policies that satisfy the constraint $d$. From this, the optimal policy is extractable with $\pi^* = \arg\underset{\pi \in \Pi_\mathcal{C}}{\max} J(\pi)$.

\subsubsection{Deep Q Learning}
Deep Q Networks (DQN) have been first introduced by \cite{mnih2013playing}. They have successfully been able to approximate control policies from high-dimensional state spaces in MDPs with the use of Artificial Neural Networks \cite{schmidhuber2015deep}. A Q~function is, for states $\mathcal{S}$ and actions $\mathcal{A}$, defined as a real-valued function $Q: \mathcal{S} \times \mathcal{A} \rightarrow \mathbb{R}$. At its core, it updates in every iteration of \term{learning} with learning rate $\alpha$, its Q~table via the Bellman update:
\begin{equation}
    Q(s_t, a_t) = (1-\alpha)Q(s_t, a_t) + \alpha (r_t + \gamma \underset{a \in \mathcal{A}}{\max} Q(s_{t+1}, a))
\end{equation}
Here, the discount factor $\gamma$ is used to assign a lower weight to future states.
Training a DQN involves sampling batches of experience that contain observations and actions taken from an environment. The network predicts Q~values and measures its loss against the target values, for the goal of minimizing said loss. Then, using stochastic gradient descent, the model gets updated.

For a stable training process, DQN uses a target network $Q'$, which is a separate neural network that is a copy of the Q~network but is updated less frequently. During training, the Q~network is used to predict the Q~values for the current state, and the target network is used to estimate the future rewards. This helps to reduce the correlation between the target and predicted Q~values, making the training process more stable and preventing divergence.

The target network is only periodically updated by copying the weights from the Q~network. This delayed update helps to ensure a more consistent and slowly moving target during the training process, making the learning process more robust and efficient.

\section{Methodology}
In this section, we propose a novel architecture for constrained Reinforcement Learning. 

 \subsection{Problem Setting Formulation}
 In classical Reinforcement Learning, state information is solely described by states $s \in \mathcal{S}$. We propose a setting, in which the agent perceives its environment through observations $\mathcal{X}$. Any ego state information of the agent, as well as action constraints, are described by $((P,T,F), \omega, \delta, M_0)$. This way, the agent is able to learn to act in a dynamic environment while being constrained by its own definition through the Petri net. We define a Reinforcement Learning Petri net as such:
 \begin{definition} \ \\
 A Reinforcement Learning Petri net (RLPN) is given by a tuple $(\mathcal{X}, r, (P, T, F), \omega, M_0, \gamma, \delta)$, where $\gamma \in [0,1]$ and $\delta \in \mathbb{N}_0$ are optional by definition.
 \end{definition}
 
 For this, we propose an extension to the classical MDP that fits the constraining nature of PNs by combining external observations and internal state information. Furthermore, we modify both the transition probability and the reward function definitions.
 \begin{definition} \ \\
     A State-Enhanced Markov Decision Process (SE-MDP) is a tuple $(\tilde{\mathcal{S}}, \mathcal{A}, \tilde{p}, \gamma, \tilde{r}, s_0)$, where $\tilde{\mathcal{S}} = \{(s,x) | s \in \mathcal{S}, x \in \mathcal{X}\}$ and $\tilde{p}: (\mathcal{S} \times \mathcal{X}) \times \mathcal{A} \times (\mathcal{S} \times \mathcal{X}) \rightarrow [0,1]$. Analogously, the reward function is defined as $\tilde{r} : (\mathcal{S} \times \mathcal{X}) \times \mathcal{A} \rightarrow \mathbb{R}$. In the constrained case, we define the SE-CMDP as $(\tilde{\mathcal{S}}, \mathcal{A}, \tilde{p}, \gamma, \tilde{r}, s_0, \mathcal{C}, d)$.
 \end{definition}

 \subsection{RLPN Architecture}

\begin{figure}[h]
    \centering
        \includegraphics[width=\linewidth]{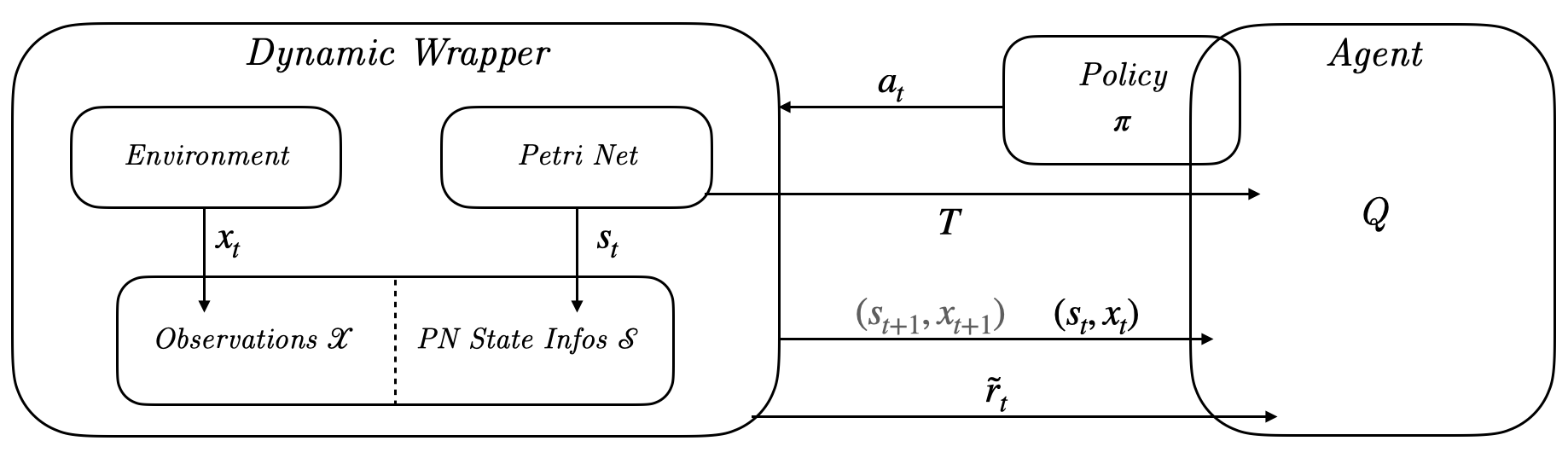}
    \caption{Proposed Constrained Reinforcement Learning closed-loop design. The dynamic wrapper contains both the environment and the Petri net and processes inputs from and outputs to the agent.}
    \label{fig:environment-setup}
\end{figure}

We highlight several advantages of our proposed architecture, which is illustrated in Figure \ref{fig:environment-setup}. 
With the requirement of needing a verified PN process model, our agent can act without any further safety-relevant constraints. As an implication, the agent can be replaced flexibly with other agents and/or control policies.

In order to limit the agent's actions based on their current state, we deem \term{Petri nets} a useful tool. As a bonus, they are widely found in business process modeling, because of their interpretable, analyzable, well-defined properties and their ability to model concurrent processes. This is an advantage over Finite State Machines (\cite{wang2019formal}), which can be considered special cases of PNs. Standard finite state machines are limited to a single state, while PNs are more complex, meaning they can have tokens in multiple locations and are able to model an exponentially larger state space.

\subsection{Constraint Enforcement}
By design, a RLPN agent is subject to constraints. Extending on work by \cite{kalweit2020deep}, we propose constraining the action directly through the update function of the Q network. 
\sloppy First, we define the set of valid actions to be
\begin{equation*}
    \mathcal{A}_\mathcal{C} ((s_t, x_t)) = \{a \in \mathcal{A} | \mathcal{C}((s_t, x_t), a_t, (s_{t+1}, x_{t+1})) \in T \}.
\end{equation*}
We enforce those constraints directly through the Q-learning update rule
\begin{equation*}
\begin{split}
    Q^\mathcal{C}((s_t, x_t), a_t) & = (1-\alpha) Q^\mathcal{C}((s_t, x_t), a_t) + \\
     & \alpha (\tilde{r} + \gamma \underset{a \in \mathcal{A}_{\mathcal{C}}((s_{t+1}, x_{t+1}))}{\max} Q^\mathcal{C}((s_{t+1}, x_{t+1}), a))
\end{split}
\end{equation*}
with learning rate $\alpha$. Since we calculate the set of actions from the PN transitions given extended state information, it is guaranteed that constraints are always adhered to.
\sloppy Finally, the optimal policy is then given by 
\begin{equation*}
    \pi^* = \arg\underset{a \in \mathcal{A}_{\mathcal{C} (s_t, x_t)}}{\max} Q^\mathcal{C}((s_t, x_t), a).
\end{equation*}
The full learning algorithm is shown in algorithm \ref{alg:pn-cdqn}.

\RestyleAlgo{ruled}
\SetKwComment{Comment}{/* }{ */}
\begin{algorithm}[hbt!]
\caption{Petri-Net-Constrained DQN}\label{alg:pn-cdqn}

    \KwData{$Q, Q', \text{replay buffer }\mathcal{B}, \text{Petri net \textit{PN}}, \text{learning rate }\alpha$}
    
    \KwResult{$y = x^n$}
    \For{rounds $t=1,\dots$}{
        sample $((s_i, x_i), a_i, (s_{i+1}, x_{i+1}), \tilde{r}_i)_{1\leq i \leq n} \in \mathcal{B}$\;
        
        calculate $\mathcal{A}_\mathcal{C}((s_{i+1}, x_{i+1}))_{1 \leq i \leq n}$ from \textit{PN} 
        
        $y_i \gets \tilde{r} + \gamma \underset{a \in \mathcal{A}_\mathcal{C}((s_{i+1}, x_{i+1}))}{\max} Q'((s_{i+1}, x_{i+1}), a | \theta^{Q'})$ 
        
        calculate loss $L_{Q^\theta} \gets \frac{1}{N}\sum\limits^N_{i=1} (Q - y_i)^2$\;
        
        take gradient step $\nabla_\theta L_{Q^\theta}$\;
        
        update target network $Q'$ weighted by $\alpha$\;
    }
\end{algorithm}

\subsection{Gymnasium Environment Wrapper}
We set up this environment to facilitate showcasing the functionality of the constraint wrapper ($CW$) around an example environment.
The wrapper is developed on top of the Gymnasium library, which emerged from OpenAI's Gym library \cite{gym}.
Its class signature is $CW(E, (P, T, F), \phi_P, \phi_T, \tilde{r})$, where $E$ represents the environment, $(P, T, F)$ are the places $P$, transitions $T$ and arcs $F$ from the \term{Petri net} and $\phi_P$ is the mapping function.
With $\phi_T$, a mapping function is defined between actions and transitions and $R$ represents a custom reward function, which can be applied.

The environment can be embedded as is and the constraints implied by a Petri net $(P, T, F)$ are automatically translated.
This is achieved by using projections between states and places, as well as transitions $T$ and the respective actions.

To keep the PN and environment state in sync, we provide the mapping function $\phi_P: P \rightarrow \mathcal{S}$ to the wrapper, which maps each place $p \in P$ from the \term{Petri net} to a state $s \in \mathcal{S}$ in the environment.
Additionally, we have a mapping function $\phi_T: T \rightarrow \mathcal{A}$ to map the transitions $\bar{t} \in T$ to actions $a \in \mathcal{A}$ to be able to check for constraint fulfillment.

Our custom reward function $\tilde{r}$ receives the old state, new state, and the chosen action as input as introduced above.
Based on this, we test the potency of PNs as a constraint mechanism for our approach in two different ways:
\begin{enumerate}
    \item The agent gets punished for invalid actions by a reward penalty.
    \item Invalid actions get learned by the DQN agent using internal information from the PN.
\end{enumerate}

\subsection{Environment Definition and Parsing}
\label{subsec:environment}

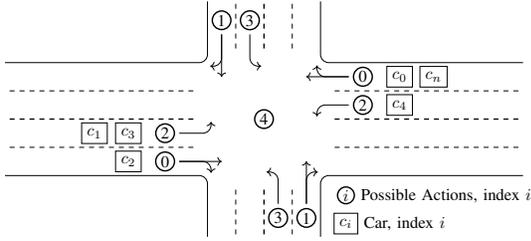
\begin{figure}[!h]
    \centering
    \resizebox{.9\columnwidth}{!}
    {
        \input{graphics/junctionScenario}
    }
    \caption{Junction scenario}
    \label{fig:junctionScenario}
\end{figure}

In this paper, we demonstrate our methods in a simple traffic junction environment in order to show the advantages of combining \term{PNs} and typical \term{RL environments}. 
The environment is specifically chosen so that you can understand the relationship between the simulation and the associated PN.
This also allows the constraints integrated in the PN for the dynamic wrapper and our PN-CDQN agent to be derived transparently.
The junction is modeled as shown in Figure \ref{fig:junctionScenario}.

The environment consists of observations of arriving and waiting cars $c_i$, where $i \in \mathcal{N}$ denotes the current car index.
Based on the car simulation in the junction environment, two observations per lane $l \in [0,7]$ are collected. First, the number of cars per lane $L_{l,0}$ is tracked, and second, the maximum waiting time of all cars in each lane $l$ ($L_{l,1}$). Here, $\textrm{lane}(c_i)$ returns the lane index of car $c_i$ is returned and $\textrm{time}(c_i)$ returns the waiting time of car $c_i$.
The values are calculated as with $|\{c_i | \textrm{lane}(c_i) = l\}|$ for $L_{l,0}$ and $\max(\{\textrm{time}(c_i) | i \in \mathbb{N}$ and $\textrm{lane}(c_i)=l \})$ for $L_{l,1}$.
In this scenario, the resulting observation space consists of 16 discrete values that represent the junction.

Additionally, the agent can choose between $9$ discrete actions, which are illustrated in Figure \ref{fig:junctionScenario} as numbered circles.
Each of the action choices $a_i$ with $i \in [0,3]$ is internally a composition of simultaneously available actions, as is the standard in traffic light cycles.
This means that choosing any \term{Green} action automatically sets the light of any other lanes to \term{Red}.
Finally, the action with index $4$ allows the agent to do nothing and thereby extend \term{Green} phases if it decides to let more cars pass.
Without any restrictions, the \term{RL agent} could allow two \term{Green} phases simultaneously. To prevent this from happening, through inherent information in the underlying PN process model, we constrain the agent accordingly.

\begin{figure}[!h]
    \centering
    \resizebox{.9\columnwidth}{!}
    {
        \input{graphics/trafficLightPetriNet}
    }
    \caption{Petri net for traffic lights constraints. North (n), East (e), South (s), and West (w) denote the lane directions.}
    \label{fig:petri-net-traffic-light}
\end{figure}
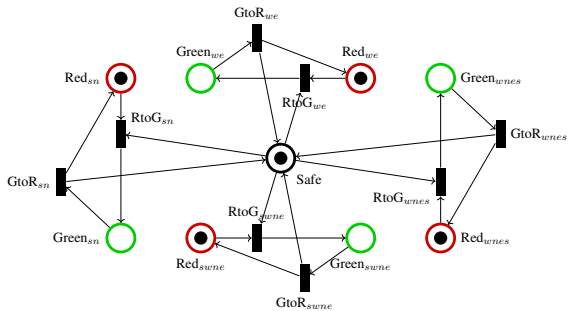

We extend the environment by building a wrapper around it.
The \term{Petri net} is constructed as shown in Figure \ref{fig:petri-net-traffic-light}.
For each lane pair (e.g. \term{North} and \term{South} left turns), it has one place for \term{Red} traffic signal state and one place for the \term{Green} state.
Transitions exist to change from \term{Green} phase to \term{Red} ($\textrm{GtoR}_{a,b}$) and inverted.
Those transitions also represent the action space, earlier described.
With the \term{Safe} place, the \term{PN} ensures to only allow one lane pair in \term{Green} state and therefore ensures, that only safe actions are executed.

\section{Experiments}

In order to demonstrate the value of our approach, in this chapter, we will present the experimental setup, as well as our results.

\subsection{Experimental Setup}
We consider the case of traffic light control in a four-way intersection, as shown in Section \ref{subsec:environment}.
In this case, for each direction \textit{(North, East, South, West)} there are two lanes consisting of a shared lane for cars driving straight and turning right, and one for cars making a left turn. We want to point out that for this paper, we chose an exemplary environment and our approach should be applicable to a variety of process control problems in other domains.

\subsection{Agent Description}
For the agent, we use the \term{Deep Q Network (DQN)} \cite{mnih2013playing} implementation from skrl (\cite{serrano2022skrl}).
The embedded \term{Q-network} and \term{Target Q-network} are deterministic Feed Forward Neural Networks, which have 2 hidden layers, each with 64 neurons \cite{bebis1994feed}.
DQN has additional hyperparameters, which we tune by building the cross-product of those and testing them prior to the experiments shown in this paper.
The learning rate is constant at $0.001$, but the random timesteps at the beginning of training, the point of the calculating the \term{Q update} function, and parameters for exploration were changed to 200,000 timesteps each.
For exploration, we leave the final exploration probability $\epsilon$ to $0.04$ but change the decrease interval to 400,000 timesteps.
The maximum amount of training timesteps is set to 15,000,000.


For the basic DQN agent, we construct a parametric reward function, with external parameters $w_i$ for $i \in [0, 4]$:
\begin{align}
    \tilde{r} = &\textrm{constraint\_fulfilled}(a) * 200 * w_0 \\
                &+max_{i \in env_{C}, j \in env_{C}}(|c_i| - |c_j|, 0) * w_1 \\
                &-sum_{l \in L}(max_{\textrm{lane}(c_x)=l, x \in env_{C}}(\textrm{time}(c_{x})) * w_2 \label{eq:waiting time}\\
                &-max_{l \in L}(max_{\textrm{lane}(c_x)=l, x \in env_{C}}(\textrm{time}(c_{x})) * w_3 \label{eq:max waiting time} \\
                &+max_{t \in \textrm{env}_{T}} (t) * w_4
\end{align}
By changing the $w_i$ factor of each summand, the influence of different environment measurements can be changed.
With $w_0$ the reward can be increased for every action, which does not violate the constraints ($\textrm{constraint\_fulfilled}(a)$ needs to be true).
The next parameter ($w_1$) increases reward, if the number of cars driven is higher, than the number of cars added. 
It also scales linear by the amount of cars driven.
Next, the lines \ref{eq:waiting time} and \ref{eq:max waiting time} are controlling the influence of cars waiting at the junction lanes.
\ref{eq:waiting time} considers all waiting times of driven cars and \ref{eq:max waiting time} only the maximum waiting time at the current state.
The last parameter $w_4$ adds a linear reward for every timestep of the simulation to reward episode completion.

As described above, we build our PN-CDQN approach on top of the base version of the DQN. As a result from the constraint enforcement, for the PN-CDQN, we are able to omit the penalty based on the boolean constraint fulfillment by setting $w_0 = 0$.

\subsection{Baseline}

To compare the performance of the wrapper in combination with both DQN and PN-CDQN, we constructed a baseline that acts similar to typical adaptive traffic control systems and therefore uses vehicle detection sensors at each lane (\textit{SCOOT}, \cite{hunt1981scoot, GARTNER1990241}).
Generally, the possibility of $k$ cars arriving at a specific time follows a nonuniform flow and is given by the \term{Poisson} distribution 
\begin{equation}
    \label{eq:poisson}
    \delta_{Poi}(k) = \frac{(\lambda t)^k \cdot e^{-\lambda t}}{k!},
\end{equation}
where $\lambda$ denotes the average arrival rate of cars per time window and $t$ is the interval duration in which cars are registered.
In each timestep, the environment has a chance to dynamically spawn cars to the lanes, and in turn, simulated presence detectors register cars and the baseline implementation of \cite{hunt1981scoot, GARTNER1990241} can be considered an adaptive round-robin cycle.
After varying the sequence of possible round-robin executions for the environment, we arrived at the sequences $B_{\textrm{v1}}$ and $B_{\textrm{v2}}$, which will be used as the baseline in the following.
Modifications of these $B_{v1}$ and $B_{v2}$ action sequences led to a deterioration in terms of performance during simulation execution.


\begin{align}
    B_{\textrm{v1}} = [&\textrm{RtoG}_{wnes}, \textrm{GtoR}_{wnes}, \textrm{RtoG}_{swne}, \textrm{GtoR}_{swne}, \label{eq:baseline-v1}\\
                     & \textrm{RtoG}_{sn}, \textrm{GtoR}_{sn}, \textrm{RtoG}_{we}, \textrm{GtoR}_{we}] \notag \\
    B_{\textrm{v2}} = [&\textrm{RtoG}_{wnes}, \textrm{GtoR}_{wnes}, \textrm{RtoG}_{swne}, \textrm{GtoR}_{swne}, \label{eq:baseline-v2} \\
                    & \textrm{RtoG}_{sn}, \textrm{DoNothing}, \textrm{GtoR}_{sn}, \textrm{RtoG}_{we}, \notag \\
                    & \textrm{DoNothing}, \textrm{GtoR}_{we}] \notag
\end{align}
As can be seen, the baselines cycle through the traffic light phases in compliance with the given Petri net transitions. After completion of a cycle, these get repeated until termination.

\subsection{Metrics}

Since we are optimizing an agent on a traffic junction environment, we need to measure the performance in this environment.
As a first and simple criterion, we will use the episode length (or: \textit{timesteps reached}) until the environment is terminated, which can be expressed as $m_{t} = max_{t \in \textrm{env}_{\textbf{T}}} (t)$, where $env_{\textbf{T}}$ contains all executed timesteps for the current simulation.

Additionally, we apply a second metric called \term{Average Junction Waiting Time (AJWT)}.
As resumed in \cite{mannion2016experimental}, this metric is one of the typically used metrics in control flow analysis that focus on traffic junctions and is therefore well suited for our considered case.
It sums up the waiting times for all cars $\textrm{time}(c_i), i \in env_{C}$ that were driven across the junction in the environment and finally divides it by the number of cars $|c_i|$, resulting in the following formula:
\begin{equation}
    m_{\textrm{ajwt}} = \frac{sum(\textrm{time}(c_i)), i \in env_{C}}{|c_i|, i \in env_{C}}
\end{equation}

\subsection{Evaluation Steps}
To evaluate the agents (DQN and PN-CDQN) and compare them with both baseline versions, we first will evaluate the optimized parameters of the reward function.
For this purpose, we will execute the training with different parameters $w_i, i \in [0, 4]$, sampled from $\{0.0, 1.0, 1.5, 2.0\}$.
By building the cross product of these parameters, we get $4^4=256$ different training executions for each model (DQN and CDQN).
In this paper, we focus on the two best performing reward parameter configurations for each respective model on timesteps reached, constraints violated and AJWT.

\subsection{Results}

We present our results in four different charts, one each  for every aspect of evaluation. 
We start with the optimization of the reward factors, continuing with performances of training models during different training phases, as well as plotting the amount of broken constraints and finally the progress of the AJWT metric over time.
We also present detailed values for each category in Figure \ref{tab:evaluation-results}.

\subsection{Influence of reward function parameters}

\begin{figure*}[!hbt]
    \centering
    \includegraphics[width=2\columnwidth]{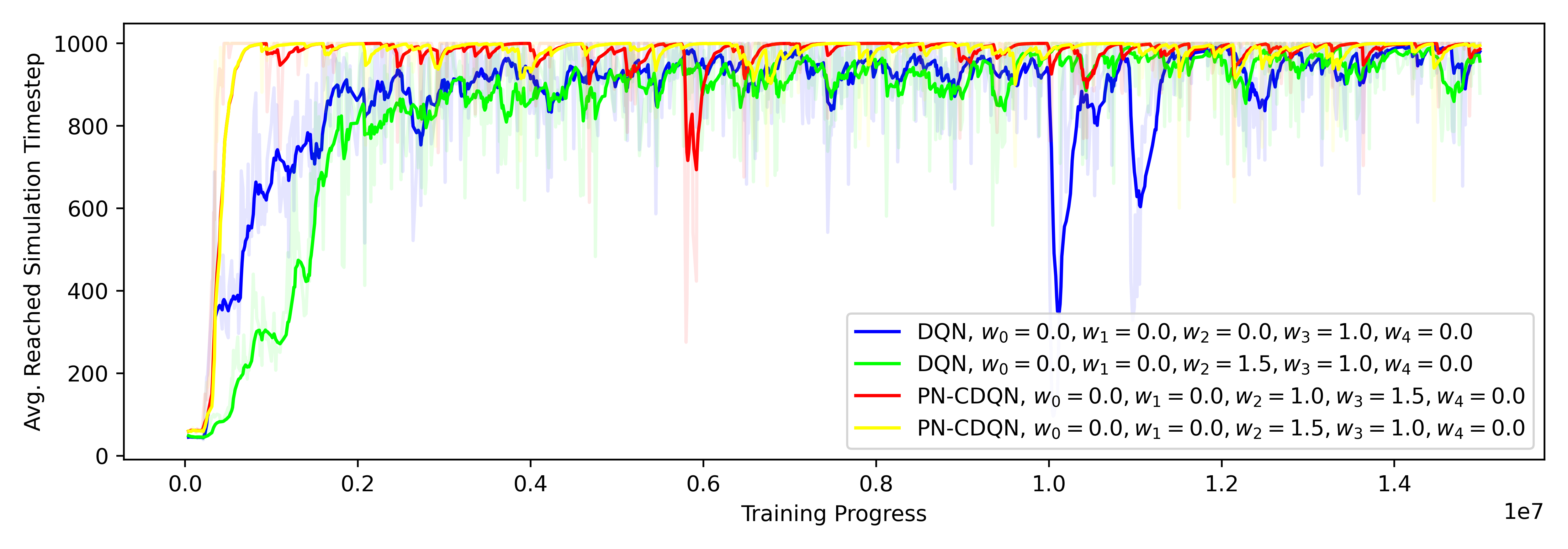}
    \caption{2 training executions for each DQN and PN-CDQN agent. Best runs, based on simulation timestep are chosen from all 256 training runs.}
    \label{fig:best-models}
\end{figure*}
The results in Figure \ref{fig:best-models} depict training runs of the two best-performing DQN and PN-CDQN agents.
For better visual comparison, we aggregate data points with a window size of $12$ into their averages. On the graph, the x-axis represents the training iterations over time and on the y-axis we plot the average simulation timestep. Here, values can vary between a minimum of $0$ and a maximum of $1000$.

We note that the best agents only use \term{waiting time} and \term{maximum waiting time} for the reward function. Additionally, the (blue) DQN agent takes has the \term{maximum waiting time} factor into consideration. All other 254 training executions, that had different reward parameters, were worse in terms of non-constraint-based performance metrics.

Besides this, we observe that during training, quickly learn within the first 4,000,000 training timesteps, an exploratory period of 200,000 training steps, as described above.
Because the PN-CDQN is implemented to only choose valid actions during exploration, the agent seems to be more efficient at exploration and thus gets a faster overview over the state space and consistently manages to reach the natural limitation of the maximum episode length.
While the DQN increases at a slower rate, it also manages to score well after training for 15,000,000 steps, at around 995 steps on average.

\subsection{Evaluation of learning progression}

\begin{figure*}[!hbt]
    \centering
    \includegraphics[width=2\columnwidth]{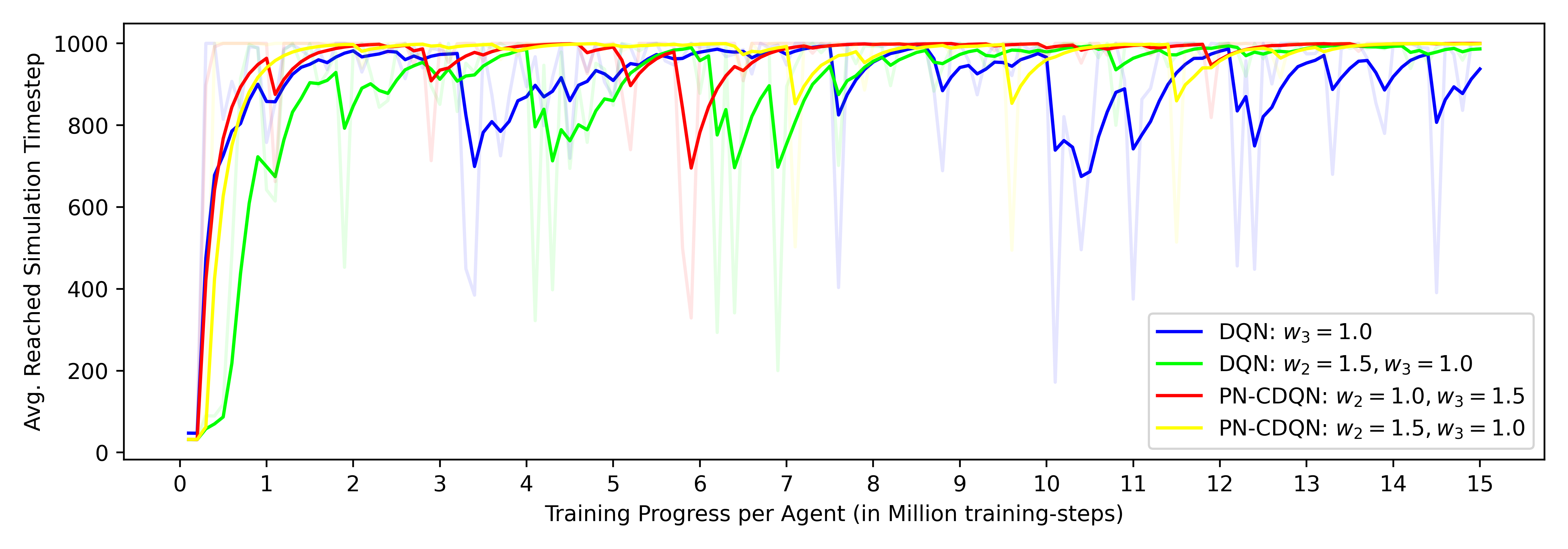}
    \caption{Evaluation results of reached simulation timesteps for intermediate trained agents. Each data point represents a training step increase of 100,000 steps. The two highest-scoring agents of both DQN and PN-CDQN are shown.}
    \label{fig:avg-timeframe}
\end{figure*}
Since Figure \ref{fig:best-models} only shows the performance during training, including the exploration phase, in order to highlight the significance of the simulation step maximization, we explicitly ran each agent 200 times on the simulation for every 100,000 training steps, resulting in Figure \ref{fig:avg-timeframe}. Here, the x-axis represents the training progress of each model.
Similar to before, graphs are smoothed to facilitate visual comparison.

When analyzing the chart, it can be seen that the plotted graphs have a similar shape as the ones in Figure \ref{fig:best-models}. The main noticeable difference is, that performance drops are more clearly visible in Figure \ref{fig:avg-timeframe}.
In general, the resulting performance for all agents (DQN and PN-CDQN) after 3,000,000 training steps is satisfactory, since agents reach the end of the simulation most of the time.

\subsection{Evaluation of violated constraints}
\begin{figure*}
    \centering
    \includegraphics[width=2\columnwidth]{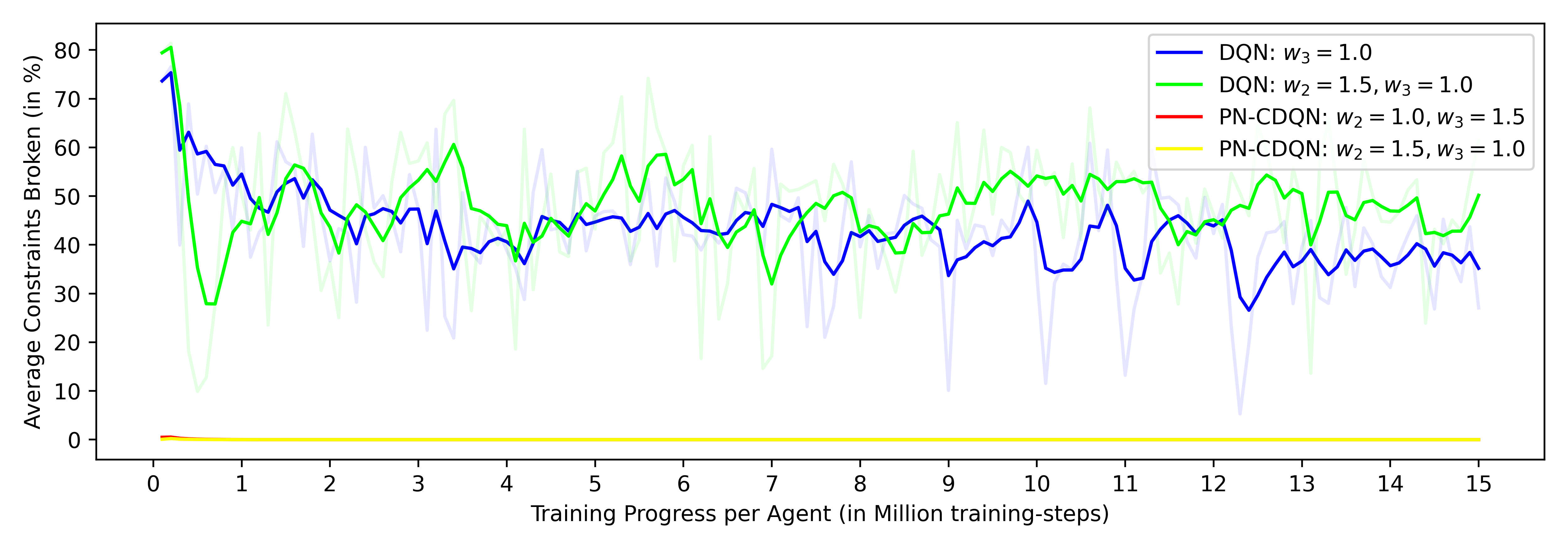}
    \caption{Evaluation results based on the amount of broken constraints for the best 2 DQN and PN-CDQN agents. Each data point evaluates an agent version, which has an increase of 100,000 training steps in comparison to the previous one.}
    \label{fig:broken-constraints}
\end{figure*}
Since we propose a novel methodology that prevents the execution of invalid actions by applying a generic wrapper during training and execution, we need to evaluate the amount of constraint violations.
Figure \ref{fig:broken-constraints} illustrates the amount of violations on hard constraints over time.
Again, each data point presented is based on an average of 200 simulations and is trained the amount of training steps, denoted at the x-axis.

When comparing the DQN and PN-CDQN agents, in this figure, it becomes evident that the PN-CDQN agent did not violate any constraints during execution.
In comparison, the DQN agent performs significantly worse. On average, 45\% of all actions it predicts are associated with a constraint violation.

However, it should be mentioned that with increasing training time, the number of broken constraints also decreases despite not having a setup that negatively rewards constraint violations.
Here, the final amount of broken constraints is around 37\% for the blue DQN agent.
It has to be noted, that all presented runs are executed with our wrapper applied and therefore, invalid actions (according to the Petri net definition) still get shielded from execution and replaced with a valid neutral action.

\subsection{Evaluation of AJWT}

\begin{figure*}[t]
    \centering
    \includegraphics[width=2\columnwidth]{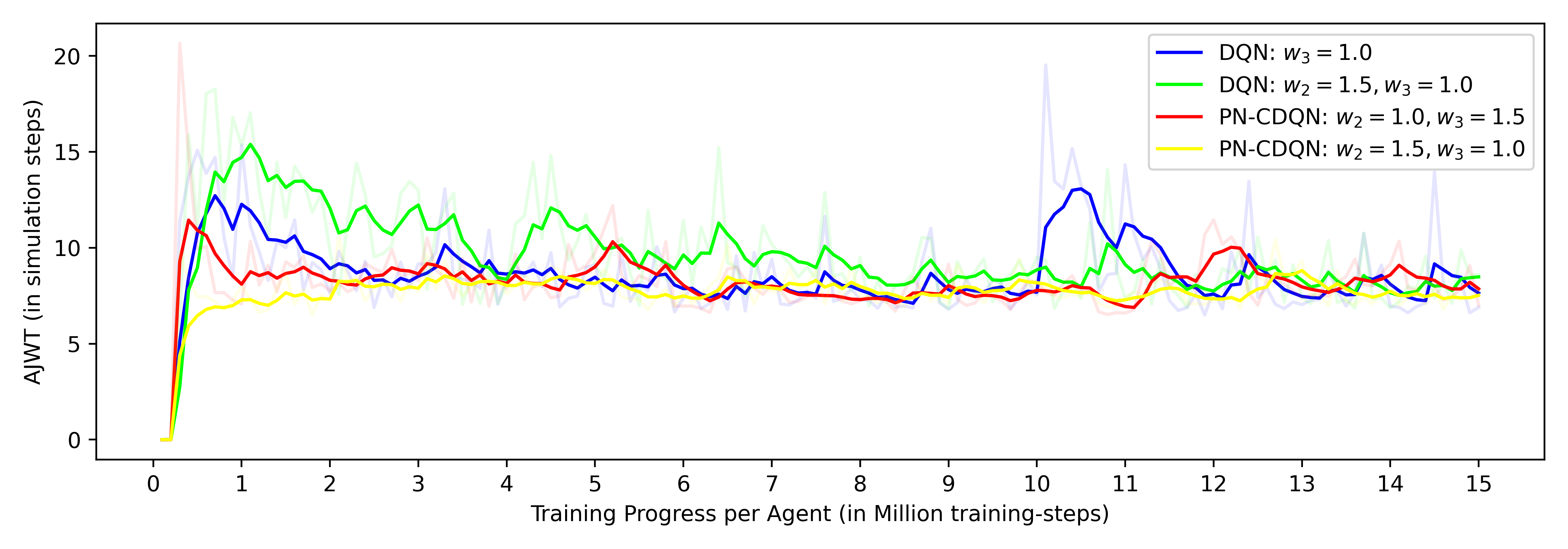}
    \caption{Evaluation results based on the AJWT metric for the best-performing DQN and PN-CDQN agents. Each data point evaluates an intermediate version of the agents in increments of 100,000 training steps.}
    \label{fig:ajwt}
\end{figure*}

Additionally to the successful results related to constraint violations, we evaluate the performance based on AJWT as metric as well, which is shown in Figure \ref{fig:ajwt}.
Again, we executed 200 simulations for each data point to negate outliers.

Given the environmental dynamics (i.e. eq.\ref{eq:poisson}) and an initially empty junction, the congestion and the resulting waiting times pick up slowly.

However, after around 200,000 training steps, the AJWT will noticeably increase and peaks at around 750,000 to 1,000,000 training steps.
Subsequently, the graph shows that the AJWT converges to a value of around 7 for all agents, with the DQN agent represented by the green line requiring significantly more training time.
Furthermore, a brief increase in AJWT is seen after 10,000,000 training steps. 

\subsection{Final evaluation at maximum episode length}


In this section, we compare the aforementioned agent configurations that were trained over the entirety of 15,000,000 training steps to the baselines $B_{v1}$ and $B_{v2}$.

\input{graphics/results/table_results_v2}

As can be seen in the results in \ref{tab:evaluation-results}, in comparison to the baselines, all trained RL agents perform much better, resulting in best \term{AJWT} values of 7.2404 (DQN) and 6.555 for the PN-CDQN agent. We note that the best PN-CDQN yields a 10.5\% performance gain over the best DQN implementations for the average, and a 3.5\% for the maxmimum, while at the same time adhering to all constraints.
The maximum value of the AJWT reflects a similar picture, where the baseline, with a value of 62.63, is at least twice as high as the worst value of the agents.
Again, our PN-CDQN approach performs better compared to the DQN agent with a value of 25.39.

The main advantage of the PN-CDQN agent becomes apparent when looking at the proportion of constraint violations, where the PN-CDQN and the baselines do not violate any hard constraints.
In contrast to this, the amount of violated constraints for the DQN agent is at least 4.5\% of all predicted actions. With an average of around 35.463\% or even 39.098\% at least every 3rd action is a constraint violation, before applying the shield from the wrapper.

In summary, the evaluation results show that problems modeled in a way that combines Petri nets and classical Reinforcement Learning are learnable.
Furthermore, it can be deduced from the results that the implementation of a generic wrapper is helpful to secure arbitrary environments using PNs, since typical agents, such as a DQN agent, can be subject to a significant amount of constraint violation.

\section{Related Work}
Several approaches in this context have been published in recent years. Learning Petri nets, as a way to build an intelligent control system combining Petri nets and RL, has been proposed by \cite{feng2010learning}. Petri nets in combination with Q~learning has been researched in the context of generating supervisory control units for production processes by~\cite{riedmann2022timed}. They conclude that basic non-deep Q~learning scales badly to complex problems, as similarity between similar Petri net states can not be measured in that instance. 

\subsection{Reinforcement Learning in Dynamic Task Scheduling}

Dynamic Task Scheduling in general is not a new topic, but based on the fact, that finding the optimal scheduling formula often is not easy, RL algorithms become more and more popular in these domains.
\cite{shyalika2020reinforcement} presented a well-described overview of current state-of-the-art methods in this domain. They provide detailed information on how model-free, as well as model-based methods can be used, to optimize scheduling tasks in RL agents.
\cite{hu2020petri} employ DRL methods to solve the dynamic scheduling problem of flexible manufacturing systems.

\subsection{Reinforcement Learning for Traffic Signal Control}

Reinforcement Learning is often applied to use cases, in which dynamic decision-making is needed.
Because it can be viewed as an infinite process, traffic signal control is a domain, which is predestined for training and running RL agents. 
In \cite{genders2016using}, the authors propose an agent that uses a method similar to DQNs to change traffic signals in a traffic simulator called Simulation of Urban Mobility (SUMO) \cite{sumo}.
Another approach was proposed in \cite{arel2010reinforcement}, where a Multi-Agent Reinforcement Learning (MARL) implementation is used to predict traffic signal control actions in 5 closely related traffic junctions.
They provide a central agent and additional outbound agents to act as information-hubs for in- and outbound traffic for the central agent. The target was to minimize delays, congestion, and the likelihood of intersection cross-blocking.

\subsection{Safe Reinforcement Learning via Shielding}

Safe RL as a topic has arisen from the need for algorithms that are supposed to perform actions dynamically on the condition of safety regarding potentially dangerous states.
Therefore, typical RL algorithms need to be limited in the choice of state-dependent actions during execution or even exploration.
Shielding as a method, to restrict actions, is proposed in \cite{alshiekh2017shielding}.
They implement a shield based on a safety automaton for a specific RL agent, which analyses the predicted action and validates if it adheres to the constraints.
If not, a safe action is returned to the environment and the agent gets the feedback of constraint violation.

\section{Conclusion}
In this paper, we proposed a way to automatically generate a constraint mechanism from a given Petri net definition for RL tasks, where the environment is partly given by the PN itself.
We showed that our proposed extension (PN-CDQN) can majorly improve performance on a given RL algorithm (DQN) over typical traffic junction controls similar to the baselines, all while complying with constraints on the environment and the agent.
It must be noted, that, while the specificly proposed PN-CDQN is based on the DQN algorithm, the overall method should be applicable to many more RL algorithms.
Since PNs are particularly useful because of their widespread use in process modeling and possible verifiability, this work aims to be an option for process model control. For this reason, our approach is not limited to traffic control but is designed to be used for a variety of tasks where the underlying process is or can be modeled with a PN.
In addition, our approach can facilitate integration of agent-based AI into these processes and increase trust by automatically constraining the agents through the generic wrapper approach.

In the future, we also plan to build upon this work to include several aspects:
First and foremost, we see great potential in this approach being extended to (decentralized) Multi-Agent Reinforcement Learning (MARL) with communication channels. With our state-enhanced MDP definition, we have both the options of connecting multiple PNs for transferring information between agents or treating shared messages as external observations. Going even further, motivated by our application-related approach, this can then also be improved by integrating \term{privacy} measures, e.g. encryption, for communication channels. We also believe that our approach can be extended to other PN types, such as Colored or Timed PNs.

\bibliographystyle{plain}
\bibliography{bibitems}

\end{document}

%% file: graphics/junctionScenario.tex
\newcommand{\distanceX}[0]{5.5}
\newcommand{\distanceY}[0]{2.5}
\newcommand{\centerDistanceX}[0]{1.5}
\newcommand{\centerDistanceY}[0]{\centerDistanceX}
\newcommand{\carLaneX}[0]{0.6}
\newcommand{\carLaneY}[0]{\carLaneX}

\begin{tikzpicture}
	\tikzset{
    	action/.style={circle,draw,line width=0.3mm,inner sep=1pt},
    	car/.style={rectangle,draw}
	}

	\draw[dashed,thick] (\centerDistanceX,0) -- (\distanceX,0);
	\draw[dashed,thick] (-\centerDistanceX,0) -- (-\distanceX,0);
	\draw[dashed,thick] (0,\centerDistanceY) -- (0,\distanceY);
	\draw[dashed,thick] (0,-\centerDistanceY) -- (0,-\distanceY);
	
	\draw[dashed] (\carLaneX,\centerDistanceY) -- (\carLaneX,\distanceY);
	\draw[dashed] (\carLaneX,-\centerDistanceY) -- (\carLaneX,-\distanceY);
	\draw[dashed] (-\carLaneX,\centerDistanceY) -- (-\carLaneX,\distanceY);
	\draw[dashed] (-\carLaneX,-\centerDistanceY) -- (-\carLaneX,-\distanceY);
	\draw[dashed] (\centerDistanceX,\carLaneY) -- (\distanceX,\carLaneY);
	\draw[dashed] (-\centerDistanceX,\carLaneY) -- (-\distanceX,\carLaneY);
	\draw[dashed] (\centerDistanceX,-\carLaneY) -- (\distanceX,-\carLaneY);
	\draw[dashed] (-\centerDistanceX,-\carLaneY) -- (-\distanceX,-\carLaneY);
	
	\draw[rounded corners=5pt] (-\distanceX,\carLaneY*2) -- (-\carLaneX*2,\carLaneY*2) -- (-\carLaneX*2,\distanceY);
	\draw[rounded corners=5pt] (-\distanceX,-\carLaneY*2) -- (-\carLaneX*2,-\carLaneY*2) -- (-\carLaneX*2,-\distanceY);\
	\draw[rounded corners=5pt] (\distanceX,\carLaneY*2) -- (\carLaneX*2,\carLaneY*2) -- (\carLaneX*2,\distanceY);
	\draw[rounded corners=5pt] (\distanceX,-\carLaneY*2) -- (\carLaneX*2,-\carLaneY*2) -- (\carLaneX*2,-\distanceY);
	
	\node[action] (E0) at (\centerDistanceX+\carLaneY,\carLaneY+\carLaneY/2) {0};
	\node[action] (E2) at (\centerDistanceX+\carLaneY,\carLaneY-\carLaneY/2) {2};
	\node[action] (W2) at (-\centerDistanceX-\carLaneY,-\carLaneY+\carLaneY/2) {2};
	\node[action] (W0) at (-\centerDistanceX-\carLaneY,-\carLaneY-\carLaneY/2) {0};
	\node[action] (S3) at (\carLaneY/2,-\centerDistanceY-\carLaneY) {3};
	\node[action] (S1) at (\carLaneY+\carLaneY/2,-\centerDistanceY-\carLaneY) {1};
	\node[action] (S3) at (-\carLaneY+\carLaneY/2,\centerDistanceY+\carLaneY) {3};
	\node[action] (S1) at (-\carLaneY-\carLaneY/2,\centerDistanceY+\carLaneY) {1};
    \node[action] (SS) at (0,0) {4};
	
	\draw[->,rounded corners=5pt] (\centerDistanceX+\carLaneY/2, \carLaneY/2) -- (\centerDistanceX*0.75,\carLaneY/2) -- (\centerDistanceX*0.75,\carLaneY/2-\carLaneY*0.4);
	\draw[->,rounded corners=5pt] (\centerDistanceX+\carLaneY/2, \carLaneY+\carLaneY/2) -- (\centerDistanceX*0.75,\carLaneY+\carLaneY/2) -- (\centerDistanceX*0.75,\carLaneY+\carLaneY/2+\carLaneY*0.4);
	\draw[->,rounded corners=5pt] (\centerDistanceX+\carLaneY/2, \carLaneY+\carLaneY/2) -- (\centerDistanceX*0.6,\carLaneY+\carLaneY/2);
	
	\draw[->,rounded corners=5pt] (-\centerDistanceX-\carLaneY/2, -\carLaneY/2) -- (-\centerDistanceX*0.75,-\carLaneY/2) -- (-\centerDistanceX*0.75,-\carLaneY/2+\carLaneY*0.4);
	\draw[->,rounded corners=5pt] (-\centerDistanceX+-\carLaneY/2, -\carLaneY-\carLaneY/2) -- (-\centerDistanceX*0.75,-\carLaneY-\carLaneY/2) -- (-\centerDistanceX*0.75,-\carLaneY-\carLaneY/2-\carLaneY*0.4);
	\draw[->,rounded corners=5pt] (-\centerDistanceX-\carLaneY/2, -\carLaneY-\carLaneY/2) -- (-\centerDistanceX*0.6,-\carLaneY-\carLaneY/2);
	
	\draw[->,rounded corners=5pt] (\carLaneX-\carLaneX/2, -\centerDistanceY-\carLaneX/2) -- (\carLaneX-\carLaneX/2,-\centerDistanceY*0.75) -- (\carLaneX-\carLaneX/2-\carLaneX*0.4,-\centerDistanceY*0.75);
	\draw[->,rounded corners=5pt] (\carLaneX+\carLaneX/2, -\centerDistanceY-\carLaneX/2) -- (\carLaneX+\carLaneX/2,-\centerDistanceY*0.75) -- (\carLaneX+\carLaneX/2+\carLaneX*0.4,-\centerDistanceY*0.75);
	\draw[->,rounded corners=5pt] (\carLaneX+\carLaneX/2, -\centerDistanceY-\carLaneX/2) -- (\carLaneX+\carLaneX/2,-\centerDistanceY*0.6);
	
	\draw[->,rounded corners=5pt] (-\carLaneX+\carLaneX/2, \centerDistanceY+\carLaneX/2) -- (-\carLaneX+\carLaneX/2,\centerDistanceY*0.75) -- (-\carLaneX+\carLaneX/2+\carLaneX*0.4,\centerDistanceY*0.75);
	\draw[->,rounded corners=5pt] (-\carLaneX-\carLaneX/2, \centerDistanceY+\carLaneX/2) -- (-\carLaneX-\carLaneX/2,\centerDistanceY*0.75) -- (-\carLaneX-\carLaneX/2-\carLaneX*0.4,\centerDistanceY*0.75);
	\draw[->,rounded corners=5pt] (-\carLaneX-\carLaneX/2, \centerDistanceY+\carLaneX/2) -- (-\carLaneX-\carLaneX/2,\centerDistanceY*0.6);
	
	\node[car] (car) at (\centerDistanceX+\carLaneY*2.3,\carLaneY+\carLaneY/2) {$c_0$};
	\node[car] (car) at (\centerDistanceX+\carLaneY*3.5,\carLaneY+\carLaneY/2) {$c_n$};
	\node[car] (car) at (\centerDistanceX+\carLaneY*2.3,\carLaneY-\carLaneY/2) {$c_4$};
	\node[car] (car) at (-\centerDistanceX-\carLaneY*2.3,-\carLaneY+\carLaneY/2) {$c_3$};
	\node[car] (car) at (-\centerDistanceX-\carLaneY*3.5,-\carLaneY+\carLaneY/2) {$c_1$};
	\node[car] (car) at (-\centerDistanceX-\carLaneY*2.3,-\carLaneY-\carLaneY/2) {$c_2$};
	
	\node[action, label=right:$\textrm{Possible Actions, index }i$] (legendAction) at (\centerDistanceX + \carLaneX*0.4, -\centerDistanceY-\carLaneY*0.2) {$i$};
	\node[car, label=right:$\textrm{Car, index }i$] (legendCar) at (\centerDistanceX + \carLaneX*0.4, -\centerDistanceY-\carLaneY*0.2-0.6) {$c_i$};
	
\end{tikzpicture}

%% file: graphics/trafficLightPetriNet.tex
\begin{tikzpicture}
	\tikzset{
    	place-red/.style = {circle,thick,draw,line width=0.7mm,minimum width=0.7cm,color=red!80!black},
    	place-green/.style = {circle,thick,draw,line width=0.7mm,minimum width=0.7cm,color=green!80!black},
    	place/.style = {circle,thick,draw,line width=0.7mm, minimum width=0.7cm},
    	transition/.style={rectangle,thick,fill, minimum width=0.1cm, minimum height=0.7cm},
    	marking/.style={circle,fill,minimum width=0.3cm}
	}
    	
    \node[place, label=below right:Safe] (Safe) at (0,0) {};
    	
    \node[place-red, label=below:{$\textrm{Red}_{swne}$}] (RedSWNE) at (-2,-2) {};
    \node[place-green, label=below:{$\textrm{Green}_{swne}$}] (GreenSWNE) at (2,-2) {};
    \node[transition, label=below:{$\textrm{GtoR}_{swne}$}] (tGtoRSWNE) at (0.6,-3) {};
    \node[transition, label=above:{$\textrm{RtoG}_{swne}$}] (tRtoGSWNE) at (-0.6,-2) {};
    	
    \node[place-green, label=above:{$\textrm{Green}_{we}$}] (GreenWE) at (-2,2) {};
    \node[place-red, label=above:{$\textrm{Red}_{we}$}] (RedWE) at (2,2) {};
    \node[transition, label=above:{$\textrm{GtoR}_{we}$}] (tGtoRWE) at (-0.6,3) {};
    \node[transition, label=below:{$\textrm{RtoG}_{we}$}] (tRtoGWE) at (0.6,2) {};
    	
    \node[place-red, label=left:{$\textrm{Red}_{sn}$}] (RedSN) at (-4,2) {};
    \node[place-green, label=left:{$\textrm{Green}_{sn}$}] (GreenSN) at(-4,-2) {};
    \node[transition, label=left:{$\textrm{GtoR}_{sn}$}] (tGtoRSN) at (-5.5,-0.6) {};
    \node[transition, label=above right:{$\textrm{RtoG}_{sn}$}] (tRtoGSN) at (-4,0.6) {};
    	
    \node[place-green, label=right:{$\textrm{Green}_{wnes}$}] (GreenWNES) at (4,2) {};
    \node[place-red, label=right:{$\textrm{Red}_{wnes}$}] (RedWNES) at (4,-2) {};
    \node[transition, label=right:{$\textrm{GtoR}_{wnes}$}] (tGtoRWNES) at (5.5,0.6) {};
    \node[transition, label=below left:{$\textrm{RtoG}_{wnes}$}] (tRtoGWNES) at (4,-0.6) {};
    
    \path [->] (RedSWNE) edge (tRtoGSWNE);
    \path [->] (tRtoGSWNE) edge (GreenSWNE);
    \path [->] (GreenSWNE) edge (tGtoRSWNE);
    \path [->] (tGtoRSWNE) edge (RedSWNE);
    \path [->] (tGtoRSWNE) edge (Safe);
    \path [->] (Safe) edge (tRtoGSWNE);
    
    \path [->] (RedWE) edge (tRtoGWE);
    \path [->] (tRtoGWE) edge (GreenWE);
    \path [->] (GreenWE) edge (tGtoRWE);
    \path [->] (tGtoRWE) edge (RedWE);
    \path [->] (tGtoRWE) edge (Safe);
    \path [->] (Safe) edge (tRtoGWE);
    
    \path [->] (RedSN) edge (tRtoGSN);
    \path [->] (tRtoGSN) edge (GreenSN);
    \path [->] (GreenSN) edge (tGtoRSN);
    \path [->] (tGtoRSN) edge (RedSN);
    \path [->] (tGtoRSN) edge (Safe);
    \path [->] (Safe) edge (tRtoGSN);
    
    \path [->] (RedWNES) edge (tRtoGWNES);
    \path [->] (tRtoGWNES) edge (GreenWNES);
    \path [->] (GreenWNES) edge (tGtoRWNES);
    \path [->] (tGtoRWNES) edge (RedWNES);
    \path [->] (tGtoRWNES) edge (Safe);
    \path [->] (Safe) edge (tRtoGWNES);
    
    \node[marking] (RedSWNE) at (-2,-2) {};
    \node[marking] (RedSN) at (-4,2) {};
    \node[marking] (RedWE) at (2,2) {};
    \node[marking] (RedWNES) at (4,-2) {};
    \node[marking] (RedWNES) at (0,0) {};
\end{tikzpicture}

%% file: graphics/results/table_results_v2.tex
\newcommand{\hl}[1]{\textbf{#1}}
\begin{table*}[ht]
\centering
\resizebox{\textwidth}{!}{
\begin{tabular}{|l|l|ccc|ccc|cc|}
\hline
\multirow{2}{*}{\textbf{Model}} & \multirow{2}{*}{\textbf{Model Parameters}} & \multicolumn{3}{c|}{\textbf{Timesteps}}    & \multicolumn{3}{c|}{\textbf{Constraints violated}} & \multicolumn{2}{c|}{\textbf{AJWT}} \\
                                     &                                            & \textbf{min} & \textbf{avg} & \textbf{max} & \textbf{min}   & \textbf{avg}   & \textbf{max}   & \textbf{avg}         & \textbf{max}        \\ \hline
\multirow{2}{*}{DQN}                 & $w_3=1.0$                       & 986          & 999.93          & \hl{1000}         & 31.700\%   & 35.463\%  & 38.200\%  & 7.2404                & 26.27              \\
                                     & $w_2=1.5$ $w_3=1.0$             & 101     & 994.54     & \hl{1000}         & 4.500\%       & 39.098\%       & 43.000\%       & 7.5622           & 30.82         \\ \hline
\multirow{2}{*}{PN-CDQN}             & $w_2=1.0$ $w_3=1.5$             & \hl{1000}           & \hl{1000}     & \hl{1000}     & \hl{0.000\%}        & \hl{0.000\%}        & \hl{0.000\%}        & \hl{6.555}               & 25.99              \\
                                     & $w_2=1.5$ $w_3=1.0$             & \hl{1000}           & \hl{1000}           & \hl{1000}          & \hl{0.000\%}        & \hl{0.000\%}        & \hl{0.000\%}        &     6.891               & \hl{25.39}              \\ \hline
\multirow{2}{*}{Baseline}            & v2                              & 243      & 395.27     & 614     & \hl{0.000\%}        & \hl{0.000\%}        & \hl{0.000\%}        & 24.508               & 112.73              \\
                                     & v1                              & 110           & 132.66           & 166           & \hl{0.000\%}        & \hl{0.000\%}        & \hl{0.000\%}        & 17.216           & 62.63        \\ \hline
\end{tabular}
}
\caption{The two best-performing DQN agents with different reward factor parameters compared to our PN-CDQN as well as the baseline versions. All agent simulations are run 200 times. For timesteps, higher values equate to a better performance, whereas for the percentage of constraint violations and the \textit{AJWT} metric, lower values are better.}
\label{tab:evaluation-results}
\end{table*}